\documentclass[letterpaper, 10 pt, conference]{ieeeconf}  

\IEEEoverridecommandlockouts                              

\usepackage{graphics} 
\usepackage{epsfig} 
\usepackage{mathptmx} 
\usepackage{times} 
\usepackage{amsmath} 
\usepackage{amssymb}  
\usepackage{tikz}
\usepackage{pifont}
\usepackage{tabularx}

\newcommand{\cmark}{\ding{51}}%
\newcommand{\xmark}{\ding{55}}%

\usepackage[hashEnumerators,smartEllipses]{markdown}

\title{\LARGE \bf
MAOMaps: A Photo-Realistic Benchmark For vSLAM and Map Merging Quality Assessment}

\author{Kirill Muraviev$^{1}$, Andrey Bokovoy$^{2}$ and Konstantin Yakovlev$^{3}$
\thanks{$^{1, 2, 3}$All authors are with the Federal Research Center for Computer Science and Control of Russian Academy of Sciences and with Moscow Institute of Physics and Technology. Konstantin Yakovlev is also with HSE University. Contact author is Kirill Muraviev
        {\tt\small muraviev@isa.ru}.}%
}

\begin{document}

\maketitle
\thispagestyle{empty}
\pagestyle{empty}

\begin{abstract}

Running numerous experiments in simulation is a necessary step before deploying a control system on a real robot. In this paper we introduce a novel benchmark that is aimed at quantitatively evaluating the quality of vision-based simultaneous localization and mapping (vSLAM) and map merging algorithms. The benchmark consists of both a dataset and a set of tools for automatic evaluation. The dataset is photo-realistic and provides both the localization and the map ground truth data. This makes it possible to evaluate not only the localization part of the SLAM pipeline but the mapping part as well. To compare the vSLAM-built maps and the ground-truth ones we introduce a novel way to find correspondences between them that takes the SLAM context into account (as opposed to other approaches like nearest neighbors). The benchmark is ROS-compatable and is open-sourced to the community.

The data and the code are available at: \texttt{github.com/CnnDepth/MAOMaps}.

\end{abstract}

\section{Introduction}

Vision-based simultaneous localization and mapping (vSLAM) problem is of the vital importance to mobile robotics since vision-sensors (RGB, RGB-D cameras) are among the most widely used ones nowadays. No wonder a wide range of different vSLAM methods and algorithms currently exist, see~\cite{milz2018visual} for an overview, and are constantly improving. Indeed, any vSLAM algorithm in the context of mobile robotics is meant to be deployed on some robotic system for some real-world application. However, before such a deployment it is reasonable to conduct a thorough empirical evaluation in simulation (i.e. not involving a mobile robot) to be able to measure the performance of the algorithm and compare it with the competitors. Thus a range of tools including benchmark datasets, performance metrics etc. are needed.

Two orthogonal requirements to a dataset which is to be used for vSLAM evaluation are as follows. First, the RGB data (i.e. the video sequence) should be photo-realistic. Second, the benchmark should contain the \emph{ground truth} data so one can measure the quality of the vSLAM algorithm(s) via some metrics. There exist datasets that are created using the data from the real robots' vision sensors, such as TUM RGB-D SLAM Dataset and Benchmark~\cite{sturm12iros} or KITTY~\cite{geiger2013vision}, and contain ground truth localization data (which is typically obtained via some external tracking system at the stage of the dataset creation). However, such datasets typically lack ground truth map data. Thus, the quantitative evaluation of the mapping part of the vSLAM method is impossible. To create a benchmark that contains the ground truth for both trajectory and map robotic simulators, such as Gazebo~\cite{koenig2004design} or Coppelia Sim~\cite{rooban2021coppeliasim}, can be used. However the RGB-data won't be realistic anymore.

\begin{figure}[t]
    \centering
    \includegraphics[width=0.45\textwidth]{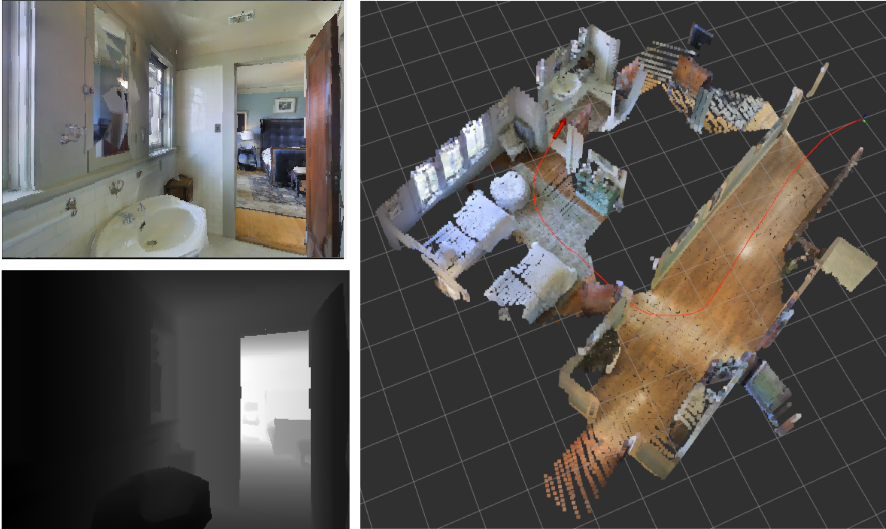}
    \caption{A sample from our dataset. Left side -- RGB image (top) and depth map (bottom) from the Habitat simulator. Right side -- ground-truth 3D map (colored pointclound) and robot's ground truth trajectory (showed as red line).}
    \label{figure_sample}
\end{figure}

In this work we suggest using a photo-realistic simulation platform Habitat~\cite{savva2019habitat}, widely used in AI research nowadays, to create a benchmark for vSLAM evaluation which contains both the trajectory and the map ground truth data on the one hand and photo-realistic RGB sequences on the other. Having access to the ground truth map data we are able to include in the presented benchmark now only the tools for the evaluation of the localization accuracy but for the map quality as well. To quantitatively measure the quality of the map reconstructed from the video data by a vSLAM algorithm we suggest finding the correspondences between the vSLAM map and the ground-truth one relying not on a trivial nearest-neighbors search but on an involved alignment procedure that considers the trajectory of a sensor as well. This contributes to getting a more fair and meaningful score that takes the context of the vSLAM problem into account.

The last but not the least, the presented dataset contains overlapping trajectories and allows to evaluate not only the vSLAM algorithms but the map-merging methods as well. Thus we name our dataset \textsc{MAOMaps} -- multi-agent overlapping maps. To the best of our knowledge it is the first dataset of such kind.

\section{Related work}


\begin{table*}[ht]
\begin{tabular}{|l|c|c|c|c|c|c|c|c|}
\hline
 & Map & Trajectory & Realistic &  Map-merge evaluation & Toolbox & \begin{tabular}[c]{@{}c@{}}Max. trajectory \\ length (m)\end{tabular} & Visual sensor & Image resolution \\ \hline
\begin{tabular}[c]{@{}l@{}}The Malaga \\ Urban Dataset\end{tabular} & \xmark  & \cmark & \cmark & \xmark & \xmark & 36 800 & Stereo & 1024x768 \\ \hline
\begin{tabular}[c]{@{}l@{}}TUM RGB-D\\ SLAM Dataset\\ and Benchmark\end{tabular} & \xmark & \cmark & \cmark & \xmark & \cmark & 40 & RGB-D & 640x480 \\ \hline
\begin{tabular}[c]{@{}l@{}}KAIST Urban \\ Dataset\end{tabular} & \xmark & \cmark & \cmark & \xmark & \xmark & 11 770 & Stereo & 1024x768 \\ \hline
EuRoC MAV & \cmark & \cmark & \cmark & \xmark & \cmark & 130 & Stereo & 752x480 \\ \hline
KITTI Dataset & \xmark & \cmark & \cmark & \xmark & \cmark & 39 200 & \begin{tabular}[c]{@{}c@{}}RGB + \\ Velodyne\end{tabular} &  1392x512 \\ \hline
\begin{tabular}[c]{@{}l@{}}Gazebo/\\ Coppelia Sim\end{tabular} & \xmark & \cmark & \xmark & \xmark & \xmark & Any & \begin{tabular}[c]{@{}c@{}}RGB \\ RGB-D\end{tabular} & Any \\ \hline
MAOMaps (ours) & \cmark & \cmark & \cmark & \cmark & \cmark & 33 & RGB-D & 320x240 \\ \hline
    \end{tabular}
    \caption{Features of different datasets and benchmarks used for vSLAM evaluation.}
    \label{tab:datasets_features}
\end{table*}


There exist a large variety of the datasets and toolboxes nowadays that are used in the community to evaluate and compare different vSLAM algorithms. We will give a brief overview here.

One of the most popular outdoor datasets is KITTI Dataset~\cite{geiger2013vision}. Is widely used in the context of autonomous driving  as it was captured by the two high-resolution color and grayscale video cameras and Velodyne laser scanner installed on top of the moving car. GPS localization data is also available. The latter can be used to compute localization accuracy of a vSLAM algorithm (the corresponding tool is included in the supplementary toolbox). Another popular datasets of that kind (outdoor scenes, autonomous driving) are Malaga Urban Data Set~\cite{blanco2013mlgdataset}, KAIST Urban Data set~\cite{jeong2019complex} and others. In all these benchmarks GPS data serves as the trajectory ground-truth in case one wants to assess the localization accuracy of a SLAM algorithm. However no tools for such assessment are usually provided by default. Map ground-truth is absent thus there is no way to access the quality of the maps built by vSLAM algorithms.


As for the indoor datasets, perhaps, the most known is the TUM RGB-D SLAM Dataset and Benchmark~\cite{sturm12iros}. It contains RGB-D image-depth pairs, captured with Microsoft Kinect. To capture some scenes the latter was installed on a moving wheeled robot (while for the others the Kinect camera was hand-held). The ground-truth localization data is provided that was reconstructed using the external motion tracking system. No ground-truth map data is available. The benchmark contains tools for color and depth images association, visualization and for the evaluation of the localization accuracy (absolute trajectory error and relative pose error).




Another indoor dataset that was captured using a real robot (micro-aerial vehicle) is  the EuRoC MAV~\cite{Burri25012016} dataset. It contains stereo images, synchronized IMU measurements, and accurate motion and structure ground-truth. The motion ground-truth data (6D poses) is constructed via the external tracking system (Vicon motion capture system, Leica MS50 Laser tracker). Map ground-truth data is represented by the detailed 3D lidar scans of the environment. EuRoC consists of 11 scenes + 1 scene for the calibration purposes. The toolbox is limited to data-loader.

Besides datasets that were captured in the real world one can use specialized robotic simulators, such as Gazebo~\cite{koenig2004design} or CopelliaSim~\cite{rooban2021coppeliasim}, to create benchmarks. The advantage of using these simulators is the ability to create as many different scenes as one wants and the access to the precise ground-truth data, being both localization data and map data. On the other hand it is hard (almost impossible) to create visually-realistic scenes in the simulators of that kind. To mitigate this issue in this work we adopt using a photo-realistic simulator Habitat. The number of environments in this dataset is limited, however one can create an infinite amount of different trajectories in those scenes. We utilize this to create overlapping trajectories that can be used for the evaluation not only of the vSLAM algorithms but of the map-merging algorithms as well. This feature is absolutely unique for our dataset (i.e. we are not aware of other datasets and benchmarks that allow evaluating map-merging algorithms).




To evaluate different vSLAM algorithms one needs not only the rich dataset but the toolbox to compute the metrics that measure the performance of the algorithms. By now the most common metrics are the ones related to the accuracy of the trajectory, e.g. absolute trajectory error (ATE) and relative pose error (RPE). Some datasets described above include tools to compute them (but this is not a very common feature).

Map quality estimation is a much more challenging task. First, very rarely the ground-truth map (in any form) is provided in the dataset. Second, even if it is available there is no common way to quantify the similarity between those maps. To the best of our knowledge, no existing vSLAM benchmark contains tools to estimate the quality of the mapping component of a vSLAM algorithm. Our benchmark provides both ground-truth map data (in the widely-used pointcloud format) and the tool to compute the accuracy of the map construction.

Table~\ref{tab:datasets_features} gives the summary of the above-said. Rows of the table correspond to different datasets used in the vSLAM community to evaluate different algorithms, columns -- to different features and specifications of those datasets.




\section{Developed benchmark}

\subsection{Dataset}

\begin{table*}[]
    \centering
    \begin{tabularx}{\textwidth}{c|c|c|c|c|c|c|c}
        Sample & 1st path length in m & 2nd path length in m & 1st bag duration in s & 2nd bag duration in s & IoU & Points 3D & Resolution 2D \\
    \hline
        1 & 21.7 & 20.3 & 54.3 & 58.3 & 0.480 & 405 288 & 399x403 \\
        2 & 32.9 & 20.8 & 66.3 & 39.4 & 0.295 & 493 070 & 448x509 \\
        3 & 13.7 & 15.9 & 28.0 & 40.8 & 0.619 & 130 162 & 234x399 \\
        4 & 19.0 & 22.9 & 41.2 & 45.4 & 0.449 & 415 810 & 463x419 \\
        5 & 18.4 & 13.8 & 30.1 & 32.2 & 0.437 & 175 628 & 383x243 \\
        6 & 16.2 & 21.2 & 29.1 & 34.3 & 0.402 & 193 740 & 444x219 \\
        7 & 4.1 & 7.8 & 23.7 & 29.7 & 0.479 & 121 247 & 239x236 \\
        8 & 13.9 & 11.0 & 31.8 & 24.7 & 0.457 & 163 418 & 329x250 \\
        9 & 9.4 & 4.3 & 23.0 & 29.6 & 0.302 & 88 337 & 211x245 \\
        10 & 9.8 & 10.0 & 28.6 & 33.1 & 0.399 & 189 478 & 250x310 \\
        11 & 21.5 & 11.2 & 46.8 & 32.8 & 0.399 & 240 041 & 330x379 \\
        12 & 16.3 & 17.4 & 36.2 & 36.0 & 0.327 & 397 487 & 528x407 \\
        13 & 22.2 & 21.0 & 39.8 & 44.0 & 0.535 & 266 472 & 329x425 \\
        14 & 13.9 & 15.8 & 20.8 & 27.8 & 0.420 & 214 593 & 327x484 \\
        15 & 12.6 & 21.1 & 31.7 & 58.7 & 0.202 & 277 755 & 397x314 \\
        16 & 21.6 & 17.8 & 44.6 & 42.0 & 0.266 & 267 133 & 400x284 \\
        17 & 10.8 & 6.5 & 23.5 & 23.8 & 0.388 & 164 773 & 349x238 \\
        18 & 13.7 & 12.3 & 28.3 & 38.2 & 0.512 & 156 416 & 255x326 \\
        19 & 15.6 & 14.1 & 30.1 & 34.8 & 0.417 & 183 705 & 228x319 \\
        20 & 18.9 & 19.9 & 34.9 & 37.9 & 0.283 & 251 173 & 293x465 \\
    \end{tabularx}
    \caption{Parameters of the samples of MAOMaps dataset.}
    \label{table_description}
\end{table*}

For comprehensive SLAM evaluation, we need both ground truth trajectory and ground truth map of the environment. Synthetic environments that can be constructed in wide-spread robotic simulators such as Gazebo usually lack details, have lightning artifacts and poor textures. Thus the RGB-sequences obtained from such simulators are non-realistic and often vSLAM algorithms perform badly on them. 
To this end, we created our dataset using the  photorealistic Habitat simulator \cite{savva2019habitat} and Matterport 3D collection \cite{chang2017matterport3d}. The latter contains a variety of photo-realistic indoor scenes and the former allows the robot to freely move in these environments. Moreover, at each time step we have an access to the exact position of the robot and the ground-truth depth map. 

For our dataset, we selected 6 scenes of Matterport 3D and performed 40 separated runs of virtual robot through these scenes in Habitat simulator. For each run we recorded ground truth trajectory, RGB images and depth maps. One of the samples from our dataset is shown in Figure~\ref{figure_sample}.

We also created precise ground truth 3D maps for each run using back-projection. For each depth $d_{h,w,t}$ on a depth map $D_t$, which was observed from position $p_t$, we find the element of the ground-truth map $m_{h,w,t} = p_t + d_{h,w,t} * r_{h,w,t}$, where $r_{h,w,t}=T*(1,\frac{w-W/2}{W/2},\frac{h-H/2}{H/2})$ and $T$ -- is the rotation operator that transforms vector $(1,0,0)$ into the camera's optical axe. We chose 0.05m. to be the discretization step for our ground-truth map.

\begin{figure}[t]
    \centering
    \includegraphics[width=0.5\textwidth]{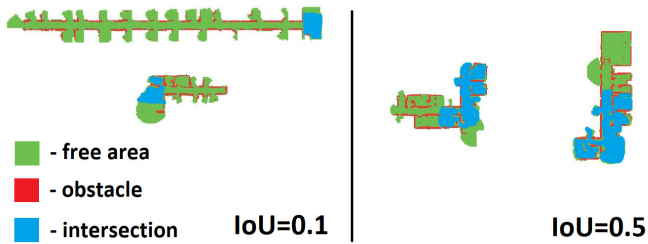}
    \caption{The IoU metric reflects the complexity of merging. Maps on the left overlap less than the maps on the right and that is captured by lower IoU (0.1 vs. 0.5). Thus it is harder to merge maps on the left.}
    \label{figure_iou}
\end{figure}

While performing the virtual runs of a robot we had in mind the map-merging scenario. Thus each even run of robot has some overlap with the previous odd run. By overlapping we mean that the robot during the even run visits some locations that were visited during the previous run. These two runs constitute a \textit{sample} in our dataset. The complexity of a map-merging task can be quantified via the \textit{intersection over union} (IoU) metric -- when IoU is bigger the merging task is obviously easier. We computed IoU for the 2D projections of the ground-truth maps -- see Figure~\ref{figure_iou}. 

Overall IoU varied from 0.20 to 0.62. Trajectory length varied from 4m to 33m, and the duration varied from 20s to 67s. Detailed information about each sample is shown in table \ref{table_description}. The last two columns show the total number of the map elements (the bigger -- the more complex the map is) and the size of the 2D projection of such map on a ground plane. Bigger sizes naturally correspond to larger maps.

To simplify vSLAM evaluation, we collected our data using Robot Operating System (ROS)\footnote{https://www.ros.org} and stored raw data for each run in Rosbag\footnote{https://wiki.ros.org/rosbag} format. The rosbags contain RGB images, depth images, camera calibration info and true robot's 6-DoF pose. These rosbags and ground truth maps and trajectories are stored in the directories called \texttt{sample1}, ..., \texttt{sample20}.
Each directory contains the following:
\begin{itemize}
    \item Raw data in Rosbag format for the first and the second run of the sample. RGB images are published to the topic named \textit{/habitat/rgb/image}, depth maps -- to \textit{/habitat/depth/image}, poses -- to  \textit{/true\_pose} and camera info -- to  \textit{/habitat/rgb/camera\_info}.
    
    \item Ground truth maps for both runs of the sample that are stored in the Pointcloud format (\text{.pcd}) and the text format as well. An example of a ground-truth pointcloud map is shown on Figure~\ref{figure_sample} on the right.
    
   \item Merged ground truth map in the Pointcloud and text format.
   
   \item Color data (RGB) for the ground truth maps.
   
   \item Ground truth trajectories (sequences of 6D poses) in text format.
   
   \item Start/goal locations of the trajectories as 6D poses.
\end{itemize}












\subsection{Toolbox}

To perform vSLAM and map merging quality estimation, we introduced a set of tools that let post-process the generated maps and trajectories and compute different metrics. 

\begin{figure}[t]
    \includegraphics[width=0.45\textwidth]{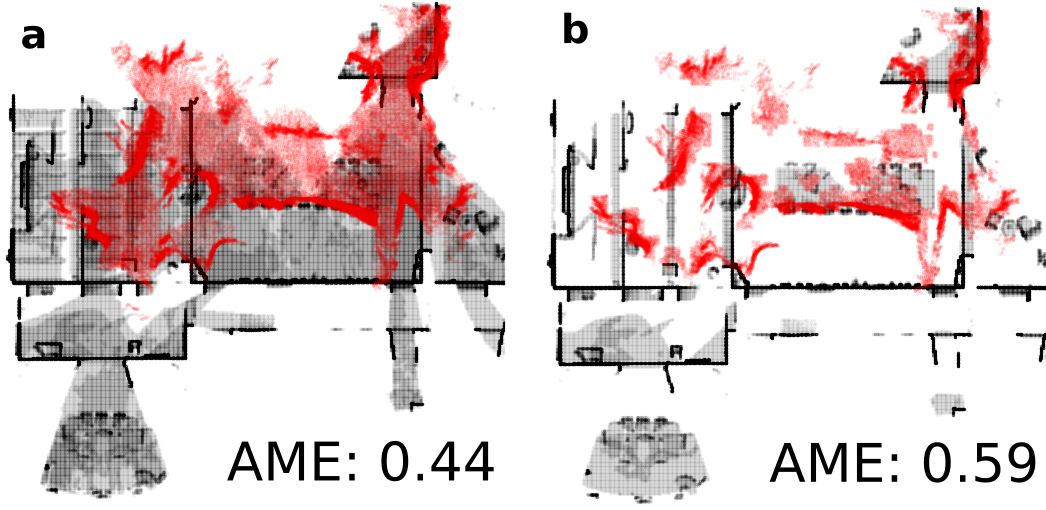}
    \caption{Original maps (a) and maps with floor removed (b). Black dots show the elements of the ground-truth map, red dots -- of the vSLAM-generated map. Map error (AME) is significantly higher when the floor is removed (0.59 vs. 0.44).}
    \label{figure_floor}
\end{figure}

Localization and mapping quality are meant to be evaluated separately. We support two conventional metrics to measure the localization accuracy -- \textbf{Absolute Trajectory Error (ATE)} and \textbf{Relative Pose Error (RPE)}:

\begin{equation}
    ATE = \sqrt{\frac{1}{T} \sum_{t=1}^T ||p_t - p_t^*||^2}
    \label{eqn_ate}
\end{equation}

\begin{equation}
    \begin{aligned}
    RPE = \sqrt{\frac{1}{T - 1} \sum_{t=1}^{T-1} ||M_{q_t}^{-1} (p_{t+1} - p_t) - (M_{q_t}^*)^{-1} (p_{t+1}^* - p_t^*)||^2}
    \end{aligned}
    \label{eqn_ate}
\end{equation}

where $p_t, p_t^*$ are ground-truth and predicted robot positions at moment $t$; $q_t, q_t^*$ are ground-truth and predicted robot orientations at moment $t$; and $M_{q_t}, M_{q_t^*}$ are rotation matrices defined by $q_t$ and $q_t^*$ respectfully.

The utility that computes these metrics can be found as \textit{collect\_data/compute\_ate\_rpe.py} in our repo.

Map quality estimation is a more involved task. In our case, when we have access to ground truth maps, it is boiled down to estimating the correspondence between the two sets of points (pointclouds) -- the ground-truth one and the constructed one. The straightforward approach for such estimation is applying the nearest neighbor method, i.e. each element (point) of the constructed map is associated with nearest point of the ground truth map. Next, by applying RMSE function to the pairs of the aligned elements, one can calculate the \textbf{Absolute Mapping Error (AME)}. This approach is implemented in CloudCompare\footnote{https://cloudcompare.org/} open source software package. So, AME metric can be computed via applying our transformation script to the SLAM-generated pointcloud (so it will be properly aligned with the ground truth one) and comparing the transformed pointcloud with the provided ground truth one in CloudCompare package.

Nearest neighbor association might be inappropriate in the context of visual SLAM. For example, when we tested RTABMAP algorithm \cite{labbe2019rtab} with CNN-generated depths \cite{bokovoy2019real} on our dataset, we noticed that the walls on the constructed map were associated with the floor. When we removed the floor from the maps (to eliminate the source of inappropriate association) AME increased by notable 1.5 times at (see Figure~\ref{figure_floor}). This case shows that a more involved association method is desirable. We have designed this method as follows.

Let the ground-truth 3D map be:    

\begin{equation}
    M=\{(x_i,y_i,z_i), i \in [1;n], x_i, y_i, z_i \in \mathbb{R}, n \in \mathbb{N}\}
    \label{eq_vslam_map}
\end{equation}

and vSLAM (predicted) map be:

\begin{equation}
    M^*=\{(x^*_i,y^*_i,z^*_i), i \in [1;N], x^*_i, y^*_i, z^*_i \in \mathbb{R}, N \in \mathbb{N}\}
    \label{eq_gt_map}
\end{equation}


Let $m^*_i \in \mathbb{R}^3$ denote a point in camera's field of view at the moment $t$; $p^*_t$ and $q^*_t$ are the predicted position and orientation of camera at $t$; $p_t$ and $q_t$ are the ground-truth position and orientation at $t$.
Lets denote $M_{q^*_t}$ and $M_{q_t}$ the rotation matrices for the quaternions $q^*_t$ and $q_t$ respectively. The vector $r_t = M^{-1}_{q^*_t}M_{q_t}(m^*_i-p^*_i)$ defines the direction towards the point $m_i$ in ground truth map. Now, the point $m^*_i$ is matched with the point $m_i = p_t + \alpha r_t$, where $\alpha$ is a minimum real number such that point $p_t + \alpha r_t$ belongs to any object in the ground-truth map. 

\begin{figure}[t!]
    \includegraphics[width=0.5\textwidth]{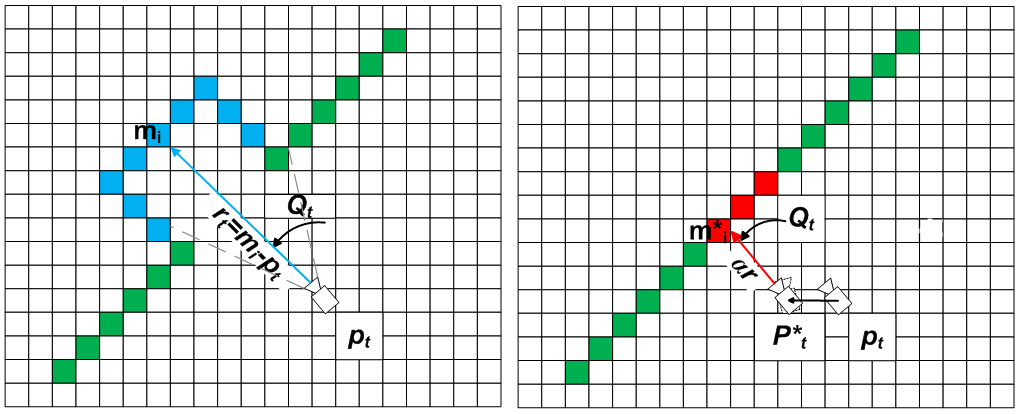}
    \caption{Our association method. Left -- the ground-truth map (2D projection). Right -- the constructed map. Map element of the constructed map $m^*_i$ is associated with the $m_i$ of the ground-truth map.}
    \label{figure_correspondences}
\end{figure}

Figure~\ref{figure_correspondences} gives an example. In plain words the idea is to transfer the ground-truth position and orientation of the camera to the predicted map and compute which point of the constructed map the camera sees from that position ($m^*_i$). This point is associated with what the camera should have seen if the vSLAM was 100\% accurate ($m^*_i$).



Overall to assess the accuracy of the map construction we suggest utilizing RMSE metric:

\begin{equation}
    AME( M, M^* )=\sqrt{\frac{1}{N} \sum^N_{t=1} || m_i^*  - f(m^*_i)||^2},
    \label{eq_ame}
\end{equation}
\noindent where $f$ is the suggested association function.

The code for our AME metric computation is integrated with Robot Operation System (ROS) and is available at \url{https://github.com/cnndepth/slam_comparison}.

\subsection{Custom dataset creator}

We have also developed a set of utilities that allows the user to extend our dataset. These tools allow creating new samples containing ground-truth maps, depths and poses. The first tool is tailored to streaming the Habitat observations (images, depths, and poses) into the ROS topics and moving virtual agent in the simulator by commands from the keyboard. 
The second tool is a set of scripts for creating ground truth maps and trajectories. It is available in our main repo.

\section{Experiments}


To evaluate the usability of our dataset and toolbox we have carried out two empirical evaluations: SLAM and map merging. In both cases only the data and the utilities from the introduced benchmark were used.

\subsection{vSLAM evaluation}
We chose RTAB-Map as the vSLAM for the evaluation as it is widely used in practice and has an open-source ROS implementation\footnote{http://wiki.ros.org/rtabmap\_ros}. Originally, RTAB-Map works with RGB-D data. We have tested it in two modes: with the RGB-D input and with the pure RGB input. In the latter case a neural network was added to the vSLAM pipeline to reconstruct depth maps from the RGB-data. The output of RTAB-Map consists of the trajectory (a sequence of 6D poses) and the map (a 3D pointcloud).

The results of experiments are presented in Table~\ref{table_slam_results}. In general, provided with the RGB-D input the algorithm is able to construct trajectories and maps with high accuracy. Average trajectory error was 0.22m, and average mapping error was 0.11m with nearest neighbor association, and 0.42m with our association procedure. We believe that the latter more accurately reflects the quality of the constructed maps.

\begin{figure}[t]
    \centering
    \includegraphics[width=0.46\textwidth]{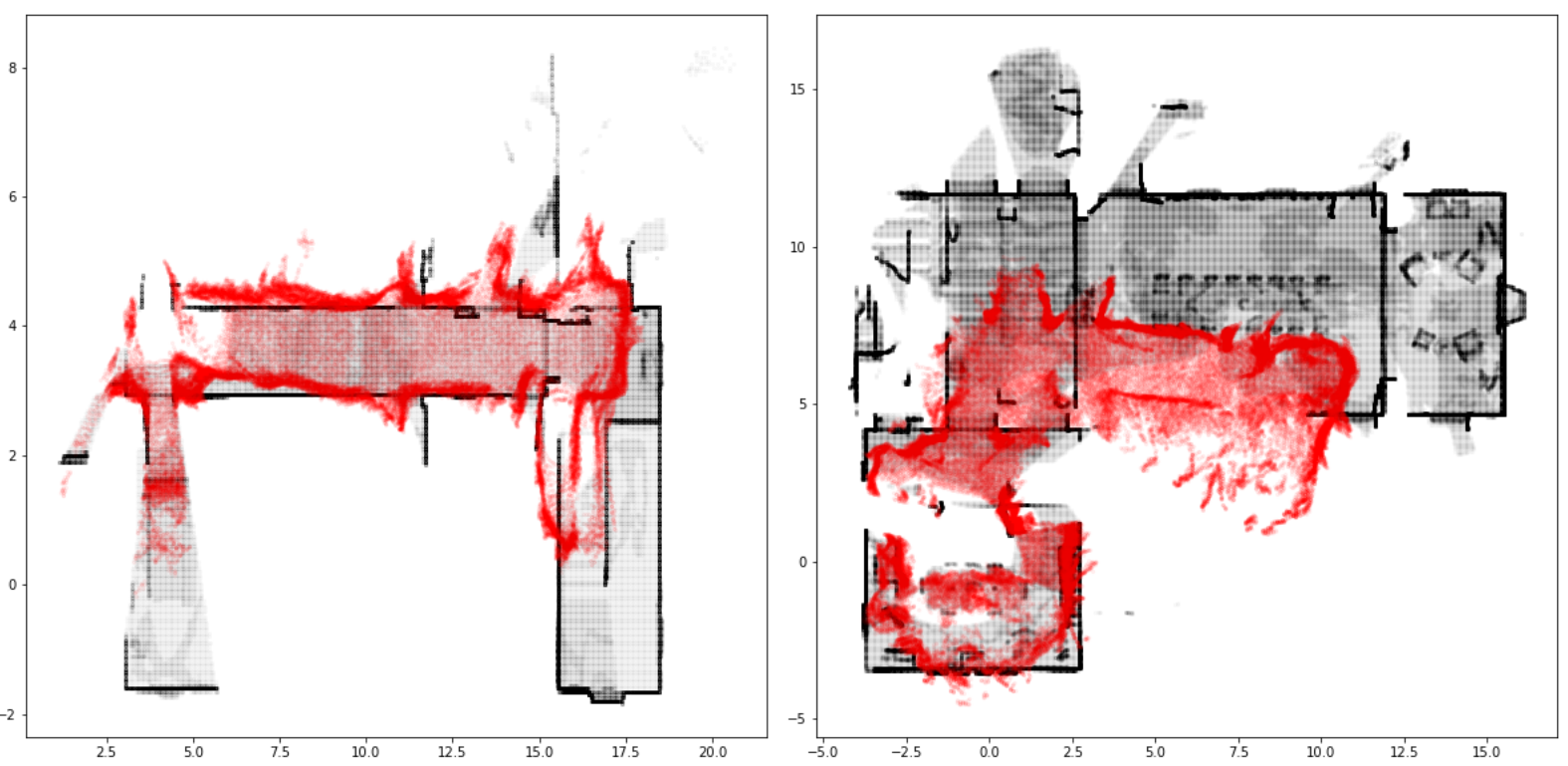}
    \caption{Examples of maps generated by RTABMAP algorithm with CNN-predicted depths. The map on left picture has relatively high quality, while map on right picture has significant deviations from ground truth. Black dots are ground truth map points, red dots - generated map points}
    \label{figure_rtabmap_fcnn}
\end{figure}

The performance of the same algorithm but with the CNN-inferred depth-maps was lower, as expected. This is adequately captured by the values of ATE, AME and AME(ours): 1.28m, 0.38m and 1.76m respectfully. This is 4 to 6 times worse compared to the results obtained from the RGB-D input. Still, the generated maps preserve the topological structure of the environments -- see Figure~\ref{figure_rtabmap_fcnn} for example. One of the most common artifacts was the scale error (as can be seen on Figure~\ref{figure_rtabmap_fcnn} on the left). This error might be caused by the fact that depth-prediction neural network was trained on another dataset, thus, after additional training on the Habitat samples the quality would possibly increase. Another common problem was the unnatural bends of the map caused by the (relatively) sharp movements of the robot when passing the corners (Figure~\ref{figure_rtabmap_fcnn} on the right). Evidently, processing such moves with RTAB-Map+CNN-depth is challenging.



\begin{table}[t]
    \centering
    \begin{tabular}{c|c|c|c}
         Method & ATE & AME & AME (ours)\\
         \hline
         RTABMAP RGB-D & 0.22 & 0.11 & 0.42\\
         RTABMAP + FCNN & 1.28 & 0.38 & 1.76\\
    \end{tabular}
    \label{table_slam_results}
    \caption{Average metrics value for vSLAM methods on our dataset samples}
\end{table}

\subsection{Map merging evaluation}

\begin{table}[t]
    \centering
    \resizebox{\columnwidth }{!}{
    \begin{tabular}{c|c|c|c|c|c|c}
        Pair & MM1 & MM2  & MM3  & MM1 AME & MM2 AME & MM3 AME \\
    \hline
    \multicolumn{2}{c}{RTAB-MAP results}\\
    \hline
        3 & + & + & + & 0.080 & 0.168 & 0.155\\
        5 & + & + & + & 0.464 & 0.127 & 0.139\\
        6 & - & + & + & x & 0.104 & 0.108\\
        7 & + & + & + & 0.063 & 0.086 & 0.076\\
        9 & - & + & - & x & 0.127 & x\\
        10 & + & - & - & 0.124 & x & x\\
        11 & + & - & - & 0.159 & x & x\\
        12 & - & + & - & x & 0.110 & x\\
        13 & + & + & + & 0.148 & 0.119 & 0.078\\
        14 & + & + & + & 0.069 & 0.118 & 0.229\\
        16 & - & + & + & x & 0.096 & 0.065\\
        18 & - & + & + & x & 0.117 & 0.103\\
        19 & - & + & - & x & 0.149 & x\\
        20 & + & - & + & 0.080 & x & 0.261\\
        
    \hline
    \multicolumn{2}{c}{RGBDSLAM\_v2 results}\\
    \hline
        
        5 & + & - & + & 0.072 & x & 0.147\\
        7 & + & + & - & 0.056 & 0.106 & x\\
        9 & + & + & - & 0.393 & 0.248 & x\\
        11 & + & + & - & 0.255 & 0.448 & x\\
        17 & - & + & - & x & 0.303 & x\\
        18 & - & + & - & x & 0.203 & x\\
        \end{tabular}
    }
    \caption{Merging results for RTAB-MAP and RGBDSLAM\_v2 algorithms with ground truth depths.}
    \label{table_rgbd}
\end{table}

In our previous work~\cite{bokovoy2020map}, we conducted a set of experiments with the open-source implementations of various map merging algorithms. We used three different implementations that were suitable for visual SLAM application on real robots: map\_merge\_2d\footnote{https://github.com/hrnr/m-explore}, map\_merging\footnote{https://github.com/emersonboyd/MultiSLAM}, and map\_merge\_3d\footnote{https://github.com/hrnr/map-merge}. The first two methods operate with 2D maps in occupancy grid format, and the last method operates with 3D pointclouds.

To get individual maps for further merging we used two different vSLAM algorithms: RTAB-Map and RGBDSLAM\_v2 \cite{endres2015robot}. Both algorithms were invoked (as before) on two different types of input: RGB-D data and RGB data with CNN-predicted depths. Each pair of the overlapping maps from our dataset was attempted to be merged by each of the three tested map merging algorithms. In case the merge was successful we measured the AME of the resultant map (with the nearest neighbor association). We didn't measure ATE and AME(ours) as there was no single trajectory in the merged map.

The best results were obtained with RTAB-Map algorithm on RGB-D input (see Table~\ref{table_rgbd}). Each of the three evaluated map merging algorithms successfully merged about a half of maps. AME of the resultant map varied from 0.06 to 0.46. These values are not much worse than AME of the individual vSLAM-constructed maps.

With RGBDSLAM algorithm, the merging results were significantly worse. The first map merging algorithm succeeded on 4 samples out of 20, the second algorithm -- on 5 samples, and the third algorithm -- only on one sample. AME for the successfully merged maps varied from 0.06 to 0.45 (this is similar to map merging of RTAB-Map's maps). Detailed results are presented in Table~\ref{table_rgbd}.

For RTAB-Map SLAM with CNN-inferred depths, each of the three map merging methods succeeded only in 10\% of runs (i.e. on 2 runs of 20) -- see Table~\ref{table_rtabmap_fcnn}). AME for the successfully merged maps varied from 0.37 to 2.21. As one can note the mapping error is significantly higher compared to when the RGB-D input was used.

For maps generated by RGBDSLAM with CNN-predicted depths, no successful merges were obtained. All of the three map merging methods failed to find the transformation between the first and the second map, because these maps were very inaccurate.

Overall, the tests of the existing map merging methods on our dataset provided a clear evidence that map merging is a very challenging problem lacking an efficient solution. Current state-of-the-art algorithms sometimes failed to merge accurate maps constructed via processing the RGB-D input. When only RGB-data was used to build maps, merging all algorithms failed in 90\% of cases.


\section{Conclusion}
We have presented a novel benchmark for the evaluation of vSLAM algorithms -- \textsc{MAOMaps}. This benchmark contains not only a photo-realistic dataset with ground truth localization and map data but also a rich tool-set which allows to measure the accuracy of both localization and mapping results of the vSLAM algorithms. Another distinctive feature of the introduced benchmark is the ability to evaluate map merging algorithms. Finally, the user is able to add samples to the dataset using the provided tools. The benchmark is ROS-compatible and open sourced. It can be found at: \texttt{github.com/CnnDepth/MAOMaps}.

\begin{table}[t]
    \centering
    \resizebox{\columnwidth }{!}{
    \begin{tabular}{c|c|c|c|c|c|c}
        Pair & MM1 & MM2 & MM3 & MM1 AME & MM2 AME & MM3 AME\\
    \hline
        3 & + & - & + & 0.821 & 0.982 & x\\
        4 & + & - & - & 0.369 & x & x\\
        6 & - & + & + & x & 1.611 & 2.209\\
        16 & - & + & - & x & 1.391 & x\\

    \end{tabular}
    }
    \caption{Merging results with RTABMAP + CNN-inferred depths.}
    \label{table_rtabmap_fcnn}
\end{table}

\addtolength{\textheight}{-12cm}   





\section*{ACKNOWLEDGMENT}

This work was supported by RSF Project \#20-71-10116. The research was carried out using the infrastructure of the shared research facilities
``High Performance Computing and Big Data'' of FRC CSC RAS (CKP ``Informatics'').

\bibliographystyle{IEEEtran}
\bibliography{IEEEabrv,references}


\end{document}